\documentclass[letterpaper, 10 pt, conference]{ieeeconf}  % Comment this line out if you need a4paper

\IEEEoverridecommandlockouts                              % This command is only needed if 
\usepackage{cite}

\usepackage{amsmath,amssymb,amsfonts, mathtools, amsthm}
\usepackage{algorithmic}
\usepackage{graphicx}
\usepackage{textcomp}
\usepackage{xcolor}
\def\BibTeX{{\rm B\kern-.05em{\sc i\kern-.025em b}\kern-.08em
    T\kern-.1667em\lower.7ex\hbox{E}\kern-.125emX}}
    
\usepackage[linesnumbered,ruled]{algorithm2e}
\usepackage{xcolor}
\usepackage{paralist}
\usepackage[caption=false]{subfig}

\usepackage{bm}
\usepackage{scalerel}

\theoremstyle{plain}
   
\theoremstyle{remark}

\theoremstyle{definition}
\newtheorem{definition}{Definition}
\theoremstyle{plain}

\newtheorem{problem}{Problem}

\DeclareMathOperator*{\argmin}{argmin} % thin space, limits underneath in displays 

\usepackage[para,online,flushleft]{threeparttable}
\usepackage{array,booktabs,makecell}
\usepackage{bm}
\let\labelindent\relax
\usepackage[inline]{enumitem}
\usepackage{booktabs}
\usepackage{relsize}

\title{\bf Risk-aware UAV-UGV Rendezvous with \\Chance-Constrained Markov Decision Process
\thanks{This work is supported in part by National Science Foundation Grant No. 1943368 and Army Grant No. W911NF2120076.}
}

\author{Guangyao Shi$^1$, Nare Karapetyan$^1$, Ahmad Bilal Asghar$^2$,  Jean-Paul Reddinger$^3$, \\James Dotterweich$^3$, James Humann$^3$, Pratap Tokekar$^1$ 
\thanks{\textsuperscript{1}University of Maryland, College Park, MD 20742 USA {\tt\small gyshi}, {\tt\small knare}, {\tt\small tokekar@umd.edu}}
\thanks{\textsuperscript{2}University of Toronto, Toronto, ON M6H 0B6 Canada {\tt\small ahmad.bilal.asghar@robotics.utias.utoronto.ca}}
\thanks{\textsuperscript{3}DEVCOM Army Research Laboratory, MD USA {\tt\small jean-paul.f.reddinger.civ}, {\tt\small james.d.humann.civ}, {\tt\small james.m.dotterweich.civ@army.mil}}
}
\begin{document}
\maketitle
\thispagestyle{empty}
\pagestyle{empty}

\begin{abstract}
We study a chance-constrained variant of the cooperative aerial-ground vehicle routing problem, in which an Unmanned Aerial Vehicle (UAV) with limited battery capacity and an Unmanned Ground Vehicle (UGV) that can also act as a mobile recharging station need to jointly accomplish a mission such as monitoring a set of points. Due to the limited battery capacity of the UAV, two vehicles sometimes have to deviate from their task to rendezvous and recharge the UAV\@. Unlike prior work that has focused on the deterministic case, we address the challenge of stochastic energy consumption of the UAV\@. We are interested in finding the optimal policy that decides when and where to rendezvous such that the expected travel time of the UAV is minimized and the probability of running out of charge is less than a user-defined tolerance.
We formulate this problem as a Chance Constrained Markov Decision Process (CCMDP). To the best knowledge of the authors, this is the first CMDP-based formulation for the UAV-UGV routing problems under power consumption uncertainty. We adopt a Linear Programming (LP) based approach to solve the problem optimally. We demonstrate the effectiveness of our formulation in the context of an Intelligence Surveillance and Reconnaissance (ISR) mission. 
\end{abstract}

\section{Introduction}

Unmanned Aerial Vehicles (UAVs) are increasingly being sought in applications such as surveillance, environmental monitoring, and agriculture due to their ability to monitor large areas in a short period of time. {One bottleneck in practice that limits their application is the limited battery capacity, especially for multi-rotor UAVs. One way to overcome this bottleneck is to use a team of aerial and ground vehicles for such tasks, in which the UGV can work as a mobile recharging station and will recharge the UAV during long-range operations. The key to achieving such cooperation on the decision-making level is to design efficient routing algorithms that can tell robots \textit{which task node to visit next}, and  \textit{when and where the UAV should be recharged}. Moreover, the rate of battery discharge of a UAV is stochastic in the real world. The routing algorithm should be able to deal with such uncertainties, e.g., trade off task performance with failure risks.} 

{In this paper, we consider the cooperative routing problem with a team of a single UGV that can work as a mobile charger and a single energy-constrained UAV, in which the UAV and UGV need to complete a task by visiting task nodes distributed throughout the task area. The UGV can only move on the road network but the can directly fly between any pair of nodes (assuming it has enough charge). Given the task nodes to visit and the stochastic energy consumption model of the UAV, we are interested in finding a routing strategy for the UAV and the UGV such that the expected time to finish the task is minimized and the probability of running out of charge less than a user-defined tolerance. Such problems can be formulated within the stochastic programming (SP) framework \cite{du2020cooperative}. However, since we need to consider not only routes but also recharge decisions and chance constraints, SP-based formulation would involve too many variables, rendering the formulation only solvable for very small instances.}

\begin{figure}
    \centering
    \includegraphics[scale=0.65]{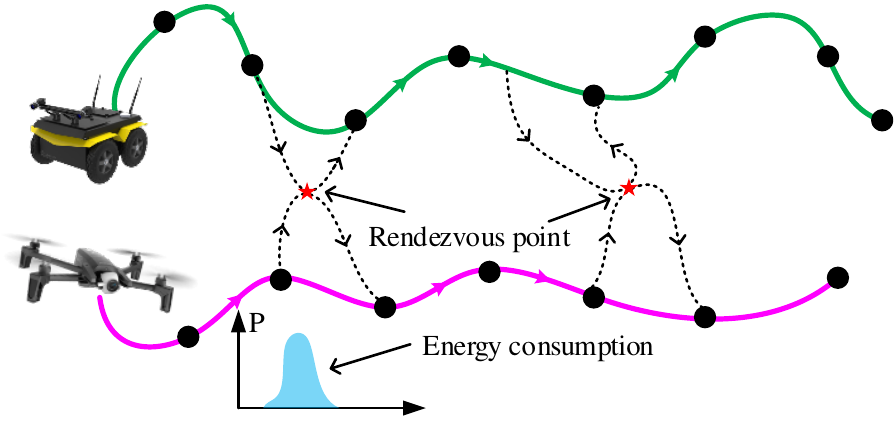}
    \caption{An illustrative example of the rendezvous problem considered in this paper. When the UAV and UGV are executing tasks, they need to decide when and where to rendezvous to replenish the battery of the UAV while minimizing the travel time of the UAV and satisfying the risk constraint induced by stochastic energy consumption. When they need to rendezvous, they will deviate from their task and meet at a chosen rendezvous location.}
    \label{fig:illustrative_example}
\end{figure}

{In this paper, we propose to find the routing strategy in two decoupled phases. In the first phase, a higher-level planner finds deterministic routes for the UAV and the UGV without considering stochasticity in energy consumption based on the task requirement. In the second phase, a risk-aware planner will refine the routes generated from the previous phase to find when and where to rendezvous to satisfy the chance constraint while minimizing the time to finish the UAV task\footnote{We focus on minimizing the time taken for the UAV task instead of the total time which would be the maximum of the UAV and UGV travel times. If the UGV task takes longer than the UAV, then once the UAV's task is done, the UGV simply executes the remaining portion of its route. Therefore, optimizing the UAV time is appropriate. Furthermore, In many applications \cite{yu2018algorithms}, the UGV's task is to simply act as a mobile recharging station}. Our focus in this paper is mainly on the second phase. We formulate our risk-aware refinement as a CMDP, in which the chance constraint is modeled as the secondary cost in the constraint. To the best of our knowledge, this is the first CMDP-based formulation for the UAV-UGV routing problems under energy consumption uncertainty. We use Linear Programming (LP) to find the optimal stationary policy. We validate our formulation and the solution in an Intelligence Surveillance and Reconnaissance (ISR) mission.}

{
The main contributions of this paper are:
\begin{itemize}
\item We show how to formulate the stochastic cooperative UAV and UGV rendezvous problem with energy constraint as a CMDP and use LP to solve the problem optimally in polynomial time. 
\item Demonstrate the effectiveness of the formulation and the solution using realistic models that are obtained using field data. 
\end{itemize}
}

The rest of this paper is organized as follows. We first give a brief overview of the related work in Section \ref{sec:rel_wrk}. Then we discuss the general problem definition along with its mathematical formulation as a Chance-Constrained Markov Decision Process (CCMDP) in Section \ref{sec:prb_frml}. Next in Section \ref{sec:sol}, we discuss the solution to this problem as an LP instance. Finally, we present the results from numerical simulations in Section \ref{section:numerical simulation} with conclusions in Section \ref{section:conclusion}.

\section{Related Work}
\label{sec:rel_wrk}

The routing of energy-constrained UAVs with stationary recharging stations or assistive UGVs has been studied extensively \cite{otto2018optimization, li2021ground, choudhury2021efficient, maini2018persistent}. Even with deterministic environmental changes or stationary conditions, this problem can be reduced to the Traveling Salesman Problem (TSP), \cite{yu2019coverage, mathew2015planning} making it an NP-hard problem. 

 The cooperative UAV and UGV routing problem has been studied from different perspectives and thus received various formulations in the literature. It is most commonly formulated as a type of vehicle routing problem. Manyam et al.~\cite{manyam2019cooperative} use a team of one UAV and one UGV with communication constraints to cooperatively visit targets. Along with an exact solver to Mixed Integer Linear Programming (MILP) formulation, they also provide heuristic reduction to the generalized traveling salesman problem (GTSP). Maini et al. \cite{maini2019cooperative} present a two-fold strategy: first, they identify feasible rendezvous points, then they formulate a MILP to find the optimal routes for the UAV and UGV. Thayer et al.~\cite{thayer2021adaptive} present a solution to the Stochastic Orienteering Problem, where the objective is to maximize the sum of rewards associated with each visited node while constrained by the maximum budget over edges with stochastic cost.

Murray and Chu~\cite{murray2015flying} introduced the flying sidekick TSP (FSTSP) for parcel delivery systems, which was later adopted in last-mile delivery applications using drones~\cite{agatz2018optimization, ha2018min}. In literature, the term multi-echelon scheme is often used for systems where delivery consists of multiple layers. Specifically, the two-echelon vehicle routing problem (2E-VRP) is concerned with finding minimal cost routes to deliver packages with trucks/UGVs and drones~\cite{li2021ground, liu2019cooperative}. An important differentiation from the original vehicle routing problem is the synchronization of UAV and UGV tasks.

Learning-based approaches have been used to address cooperative UAV and UGV routing problems. Ermugan et al.~\cite{ermaugan2022learning} also propose a two-phase approach. First, they find a route for UAVs without taking into account the energy constraints. Then, the planner learns to insert into the route recharging stations and replans a new TSP route. Reinforcement learning has also proven to be a possible approach to solving this problem \cite{wu2021reinforcement}.

In our previous work, we studied cooperative planning with a single UGV and an energy-constrained UAV as well~\cite{tokekar2016sensor, yu2018algorithms, yu2019coverage}.  Our proposed approach in~\cite{tokekar2016sensor}  demonstrated how to maximize the number of sites visited in a single charge in conjunction with the ability to land a UAV on top of a UGV to be transported to the next take-off site. We extended this in~\cite{yu2018algorithms} to allow the UAV to also be recharged while either being transported or stationary on the UGV\@.  We extended the latter to the area coverage path planning problem by formulating it as a GTSP~\cite{yu2019coverage}. Here, we extend this body of work by introducing the stochasticity of the UAV’s energy consumption and by assuming that the UGV has its own required set of tasks to be carried out. To the best of our knowledge and based on the presented literature review, none of the works takes into account the stochastic nature of energy consumption.

\section{Problem Formulation}
\label{sec:prb_frml}

The cooperative routing problem studied in this paper involves one UAV and one UGV\@. The UAV and the UGV are executing  tasks, which are given by some task planners as shown in Section \ref{subsection: UAV_task_planning}. The UAV needs to visit a sequence of task nodes in order to finish the task, but its battery may not be enough for it to finish the task in a single flight without recharging. Also, the energy consumption of the UAV is stochastic. The UAV needs to decide \textit{when} and \textit{where} it should rendezvous with the UGV to replenish the battery while minimizing the total travel time to finish the task. When the UAV decides to rendezvous with the UGV to replenish power, both the UAV and the UGV will take a detour from their respective tasks and go back to their tasks after recharging.   

At a high level, the problem studied in this paper is stated below.
\begin{problem}[Risk-aware UAV-UGV rendezvous]\label{problem:risk-aware-rendezvous}
Given a route of nodes for the UAV $\mathcal{T}_A$, a route of nodes for the UGV $\mathcal{T}_G$, and the stochastic energy consumption model and battery capacity of the UAV, find a policy for the UAV to decide when and where to rendezvous with the UGV for recharging such that the total travel time is minimized and the probability of running out of charge during flight is less than a given tolerance. 
\end{problem}
Next, we will present the setup and the assumptions that we use in this paper. Then we will present our CCMDP based formulation and show how to transform a  CCMDP into a CMDP\@. 
\subsection{Environment and Task Model}
Our problem considers a two-dimensional Euclidean space, which 
consists of a road network graph $G=(V_r, E)$ and a set of task points $V_t$ for the UAV to visit. 

The UGV has to move on the road network and its task is specified as a sequence of nodes of the road network. UAV's task is specified by some task planners using nodes in $V_t$. More details on the task planner will be discussed in the Section \ref{subsection: UAV_task_planning}. Both UAV and UGV should follow the task specification to visit the task nodes in order and they will deviate from the task route to rendezvous when necessary. 
\subsection{Vehicle Motion Model}
The UGV will move at a fixed speed $v_g$ when it transits between two nodes in the road network. When the UAV transits between two nodes, it will fly with either the best endurance speed, $v_{be}$, or the best range speed, $v_{br}$. The best endurance speed is the speed at which the energy consumption rate is minimized. At this speed, the propellers of the multirotor operate more efficiently than in hover, and the UAV is capable of the greatest flight duration. By contrast, when a UAV flies at the best range speed, it minimizes the derivative of energy consumption rate with respect to velocity. This flight speed results in a lower flight duration than operation at $v_{be}$, but will allow a greater range to be traveled per unit of energy. For a no-wind condition, the velocity of the best range is always better than the velocity of best endurance. 

\subsection{Recharging and Stochastic Energy Consumption Model}
We assume that it takes constant time $T$ to finish the recharging process, which includes the landing/take-off time and battery-swapping time. 

In this paper, we only consider the power consumption when a UAV traverses the route with the assumption that the power needed for computation, takeoff, and landing has been reserved by the power management system. As described in the transition model, the UAV will fly at a fixed speed when it transits between two nodes in the environment. However, given that constant speed, the energy consumption is stochastic considering the disturbances in the environment. 
\iffalse
One example is 
\begin{equation*}
    P_{\bm{\Theta}}(v)=\bm{\xi}_0 + \bm{\xi}_1 v + \bm{\xi}_2 v^2 + \bm{\xi}_3 v^3,
\end{equation*}
where $\bm{\Theta}=[\bm{\xi}_0, \bm{\xi}_1, \bm{\xi}_2, \bm{\xi}_3]$.
\fi

Given the distance $l$ between two task nodes and the flying speed $v$, the energy consumption can be computed as
\begin{equation}\label{equation:energy_consumption}
    e_{l, v} = \int_{t=0}^{\frac{l}{v}} P_{\bm{\Theta}}(v) dt,
\end{equation}
where  $P_{\bm{\Theta}}(v)$ is the power consumption of the UAV when it flies at a speed $v$, and $\bm{\Theta}$ is a vector of parameters for stochastic variables. 
\subsection{Rendezvous Model}
\begin{figure}
    \centering
    \includegraphics[scale=0.75]{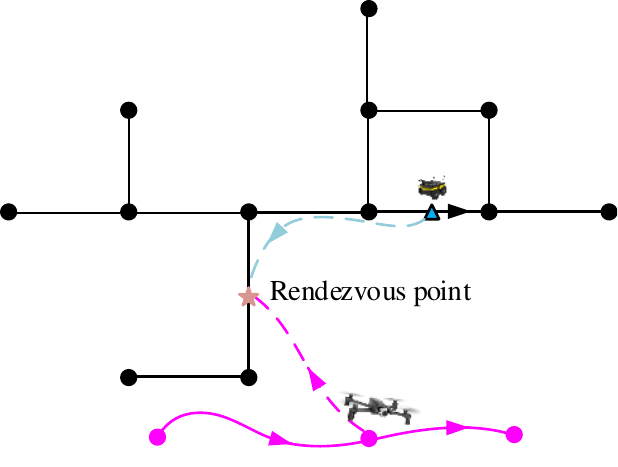}
    \caption{First step in the rendezvous process. The UGV (blue triangle) needs to deviate from its task node to rendezvous with the UAV at the rendezvous point (pink star). The rendezvous paths are in dashed lines.}
    \label{fig:rendezvous_process}
\end{figure}
\iffalse
\subsubsection{No communication}
When the UAV reaches a node and decides to rendezvous with the UGV, it will not send a rendezvous request to UGV. Instead, it will decide the rendezvous position and time based on the deterministic motion model of the UGV\@. Let $\mathcal{P}_g: T \to \mathbb{R}^2$ be the path executed by the UGV and $d: \mathbb{R}^2 \times \mathbb{R}^2 \to \mathbb{R}_{+}$ be the distance metric between two points. We use $p_a(t_0) \in \mathbb{R}^2$ and $p_g(t_0) \in \mathbb{R}^2$ to denote the position of the UAV and UGV when UAV decides to rendezvous at time $t$. The task node after $p_a(t)$ for the UAV is denoted as $p_a(t^{+})$. Then problem to find the rendezvous point can be formulated as 
\begin{align}
    \min_{p_r \in \mathcal{P}_g(t:\rm{end})} & \Delta t + \frac{d(p_r, p_a(t^{+}))}{v_a} \\
    \rm{s.t.}~& p_g(t_0 + \Delta t) = p_r \\
    & \frac{d(p_a(t_0), p_r)}{v_a} \leq \Delta t
\end{align}
\fi
When the UAV reaches a node in $\mathcal{T}_A$ and decides to rendezvous with the UGV, the UAV and UGV will deviate from their task temporarily to finish the rendezvous process. There are two steps in the rendezvous process. In the first step, the UAV and UGV will meet at a rendezvous point as shown in Fig. \ref{fig:rendezvous_process} and in the second step, they will go to the next task node in $\mathcal{T}_a$ and $\mathcal{T}_g$ respectively. We want to optimize the time consumed in these two steps to find the optimal rendezvous points.

Let $d: \mathbb{R}^2 \times \mathbb{R}^2 \to \mathbb{R}_{+}$ be the distance metric between two points in the Euclidean space. We use $d_G: V_r \times V_r \to \mathbb{R}_{+}$ to denote the length of the shortest path between two nodes in the road network. We use $\mathcal{T}_a(k)$ to denote the position of the UAV when it decides to rendezvous at  the  $k$th node in its task route and $\mathcal{T}_a(k+1)$ to denote the next position to visit for UAV after the rendezvous. With a slight abuse of notation, we use $\mathcal{T}_g(k)$ to denote the position of the UGV in the road network when the UAV decides to rendezvous at  the  $k$th node in its task route. With the above notations, the problem to find the rendezvous point can be stated below.
\begin{problem}[Where to rendezvous]\label{problem:where_to_rendezvous}
Given the positions of UAV ($\mathcal{T}_a(k)$) and UGV ($\mathcal{T}_g(k)$) at the beginning of the rendezvous process, UAV's next position to go $\mathcal{T}_a(k+1)$, UAV's flight speed $v_a$, UGV's transition speed $v_g$, and the road network $G$, we want to find a rendezvous point $p_r \in G$ such that the time consumed in the rendezvous process is minimized. Mathematically, 
\begin{align}
    & \min_{\Delta \geq 0, ~p_r \in G}  ~ \Delta + \frac{d(p_r, ~\mathcal{T}_a(k+1))}{v_a} \\
    &\text{s.t.}~ \Delta = \max (\frac{d_G(\mathcal{T}_g(k), ~p_r)}{v_g}, ~\frac{d(\mathcal{T}_a(k), ~p_r)}{v_a}).
    \iffalse
    %\rm{s.t.}~&  \frac{d_G(\mathcal{T}_g(k), ~p_r)}{v_g} \leq \Delta  \label{problem: rendezvous_point_ugv}\\
    %& \frac{d(\mathcal{T}_a(k), ~p_r)}{v_a} \leq \Delta \label{problem: rendezvous_point_uav}.
    \fi
\end{align}
\end{problem}

In the first step of the rendezvous process,  if the UAV or UGV reaches the rendezvous first, it has to wait for the other vehicle. Therefore, the time consumed in the first step is decided by the vehicle that reaches the rendezvous point later than the other. We encode this fact in the optimization problem by introducing the variable $\Delta$, which describes the maximum time needed for both UAV and UGV to reach the rendezvous point. The time consumed in the second step of the rendezvous process is the time needed for the UAV to fly back to its next task node. 

Problem \ref{problem:where_to_rendezvous} can be solved by iterating over the nodes in the road network as what we do in the case study. But such a method will increase the time to extract transition information for the CMDP\@. More efficient way to solving Problem \ref{problem:where_to_rendezvous} is left for our future work.

\subsection{Chance-Constrained Markov Decision Process}
\begin{figure}
    \centering
    \includegraphics[scale=0.65]{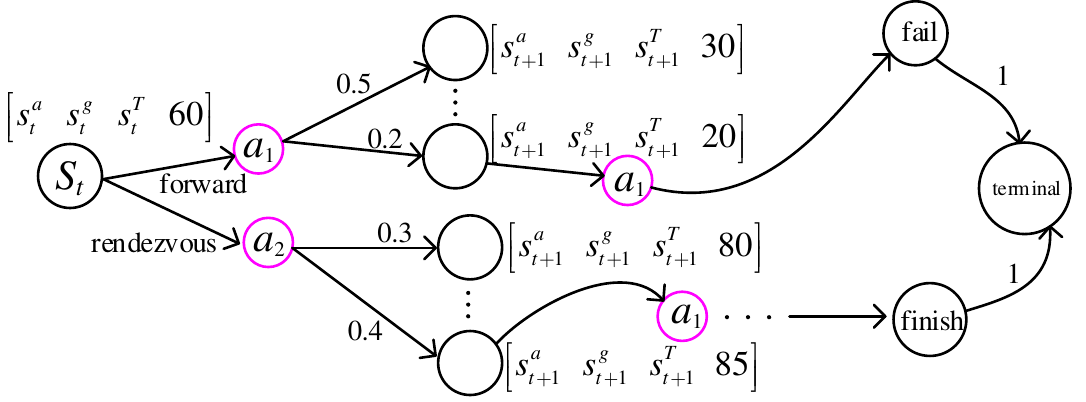}
    \caption{State transition graph in CMDP.}
    \label{fig:state_transition}
\end{figure}
One natural choice to model the sequential decision-making problems described in Problem \ref{problem:risk-aware-rendezvous} is to use MDP\@. In this section, we first show how to formulate Problem \ref{problem:risk-aware-rendezvous} as a CCMDP and then show how to transform a CCMDP into a CMDP in the following section.

The rigorous definition of an MDP can be found in \cite{bertsekas2012dynamic}. Here we define the MDP from the perspective of the application. The MDP corresponding to  Problem \ref{problem:risk-aware-rendezvous} is defined as a tuple $\mathcal{M} = (S, A, T, C,  s_0)$, where 
\begin{itemize}[leftmargin=*]
    \item $S = \mathcal{T}_a \times S_g \times \mathcal{T}_g \times \mathcal{B} ~\cup~ \{s_{ob}, ~s_l\}$ is the state space of the problem, where $\mathcal{T}_a$ here is used as an un-ordered set, which describes all possible positions of UAV in a task route; $S_g$ is the set of positions of UGV and this information is needed when we compute the rendezvous points; $\mathcal{T}_g$ here is used as an un-ordered set, which describes the task nodes UGV will visit.
    $\mathcal{T}_g$ is included in the state space  to inform the MDP about the next node the UGV needs to visit after a rendezvous. Without this information, the system will be non-Markovian;  $\mathcal{B}$ is a discretized variable for describing the state of the charge of the UAV;  $s_{ob}$ is one failure state representing the out-of-charge state and the UAV will transit to this state whenever it cannot finish its task route;  $s_{l}$ is added as an absorbing state and UAV will transit to this state when it either finishes UAV's route or runs into a failure state. {One illustrative example of state transitions is given in Fig. \ref{fig:state_transition}.}
      \ item $A$ is the action space of the UAV\@. If the UAV has not finished its route and is not in a failure state, there are four actions for the UAV to choose: 
 \begin{enumerate*}
        \item $v_{be}$: move to the next node in $\mathcal{T}_a$ with the best endurance velocity. 
        \item $v_{br}$: move to the next node in $\mathcal{T}_a$ with the best range velocity. 
        \item $v_{be\_be}$: rendezvous with the best endurance velocity. 
        \item $v_{br\_br}$: rendezvous with the best range velocity.
    \end{enumerate*}
    When UAV is in a failure state or has finished its route, there is only one action that makes the system transit to the terminal state $s_l$.
    \item $T_a(s, s^{\prime}, a) = P( s^{\prime} \mid s, a)$ is the transition function, which depends on  the stochastic energy consumption model. When UAV chooses to move forward to its next task node, its battery state at the destination node is a random variable that depends on the current battery state and Equation (1). Since we have discretized the battery charging levels at each node, the probability of reaching the destination node with a given battery charge can be calculated using Equation \eqref{equation:energy_consumption}. When it cannot reach the next task node, i.e., with non-zero probability, it will run out of charge, it transits to the failure state $s_{ob}$. When UAV chooses to rendezvous,  a rendezvous point is first computed by solving the Problem \ref{problem:where_to_rendezvous}. Then the distribution of battery remaining when it reaches the rendezvous point can be computed based on Equation \eqref{equation:energy_consumption}. The non-positive portion of the distribution corresponds to the failure probability. After recharging, UAV will transit to its next task node starting with a full battery. 
    When the UAV transits to the failure state or it finishes the task route, it will transit to the terminal state $s_l$ with probability 1 as shown in Fig. \ref{fig:state_transition}. In the terminal state $s_l$, the system will loop over this state.
    \item $C(s, s^{\prime}, a)$ is the cost function for the UAV\@. We define it as the time needed to transit between two states. If the UAV chooses to move to the next node, the cost will be time consumed during that transition. If a UAV chooses to rendezvous, the cost will be the sum of the time consumed in two steps of the rendezvous process. When the state transits to the failure state or to the terminal state, it takes zero cost.
    \item $s_0$ is the initial state of the system.
\end{itemize}

\begin{definition}[Risk]
Let $\pi$ be a policy, the risk of the policy given initial state $s_0$ is defined as 
\begin{equation}
    \rho^{\pi} (s_0) = \mathbf{P}(\exists ~t ~s_t = s_{ob} \mid s_0).
\end{equation}
\end{definition}

We seek the optimal policy $\pi^*$ that satisfies
\begin{align}
    \pi^* = & \argmin_{\pi} \mathbb{E}\left [ \sum_{i=0}^{\infty} C(s_i, \pi(s_i)) \right ]  \\
    \rm{s.t.}~&~ \rho^{\pi}(s_0) \leq \delta,
\end{align}
where $\delta$ is the user-specified risk tolerance. 
\subsection{Constrained Markov Decision Process}

We can transform a CCMDP into a CMDP by introducing a new cost function $\overline{C}: S \times S \times A \to \{0, 1\}$ \cite{geibel2005risk}. As shown in Fig. \ref{fig:state_transition}, when the system transits from a non-failure state to the failure state $s_{ob}$, it will incur a cost of one and other transitions will incur zero cost. The new cost function $\overline{C}$ is defined as 
\begin{equation}
    \overline{C}(s, a, s^{\prime}) =
    \begin{cases*}
       1  & \rm{if}~$s \neq s_{ob}$ and $s^{\prime}=s_{ob}$ \\
       0  & \rm{else}.
    \end{cases*}
\end{equation}

As shown in \cite{geibel2005risk} [Proposition 4.1],  the risk can be defined using the new cost function $\overline{C}$ as 
\begin{equation}
    \rho^{\pi}(s_0) = \mathbb{E} \left [    \sum_{i=0}^{\infty}  \overline{C}(S_i, \pi(S_i)) \mid s_0 \right].
\end{equation}
As a result, the CCMDP problem can be formulated as 
\begin{align}
    \pi^* = & \argmin_{\pi} \mathbb{E}\left [ \sum_{i=0}^{\infty} C(s_i, \pi(s_i)) \right ]  \\
    \rm{s.t.}~&~ \mathbb{E} \left [    \sum_{i=0}^{\infty}  \overline{C}(S_i, \pi(S_i)) \mid s_0 \right] \leq \delta.
\end{align}

\section{Solutions to CMDP}
\label{sec:sol}

A CMDP can be solved using Linear Programming (LP) \cite{altman1999constrained, thiebaux2016rao}. The decision variables $y$ in LP are the occupancy measure for each state-action pair and are defined as
\begin{equation}
    y(s, a) = \sum_{t} \text{Pr} (S_t = s, A_t = a).
\end{equation}
The LP is formulated as:
\begin{align}
    & \min_{y(s, a), \forall s, a}  \sum_{(s, a) \in S \times A} y(s, a) C(s, a) \\
   \rm{s.t.}~ & \sum_{s, a} y(s, a) \overline{C}(s, a) \leq \delta \label{problem:LP_cost_constraint}\\  
   &\begin{aligned}
    \sum_{a^{\prime}} y(s^{\prime}, a^{\prime}) = \mathbb{I}(s^{\prime}, s_0)+\sum_{s, a} y(s, a) \rm{Pr}(s^{\prime} \mid s, a)~\\
    \forall s^{\prime} \in S \setminus \{s_l\}
   \end{aligned}\label{problem:LP_flow_constraint}\\
   & y(s, a) \geq 0 ~ \forall s, a \label{problem:LP_occupancy_measure_constraint}, 
\end{align}
where $\mathbb{I}(s^{\prime}, s_0)$ is a Dirac delta function that returns 1 when $s^{\prime}=s_0$ and 0 otherwise. This LP corresponds to the dual linear program for MDPs \cite{altman1999constrained} with one extra cost constraint \eqref{problem:LP_cost_constraint}, which enforces that the cost of entering the failure state be lower than the predefined risk tolerance. Constraint \eqref{problem:LP_flow_constraint} is a flow conservation constraint to define valid occupancy measures and is defined by the initial state and the transition probability (see \cite{altman1999constrained}, ch. 8 for details). The last constraint \eqref{problem:LP_occupancy_measure_constraint} is added to guarantee that $y(s, a)$ is non-negative.

If LP admits a solution, we can construct the policy from the occupancy measures by normalizing them:
\begin{equation}\label{policy_traction}
    \pi^*(s, a) = \frac{y(s, a)}{\sum_{a^{\prime}} y(s, a^{\prime})}~ \forall (s, a) \in S \times A,
\end{equation}
where $\pi^*(s, a)$ is the probability of taking action $a$ in the state $s$ in the optimal stationary randomized policy. 
If Eq. \eqref{policy_traction} has a zero denominator, which suggests that state $s$ is not reachable from $s_0$, the policy for $(s, a)$ can be defined arbitrarily. 

\begin{figure}
    \centering
    \includegraphics[width=0.38\textwidth]{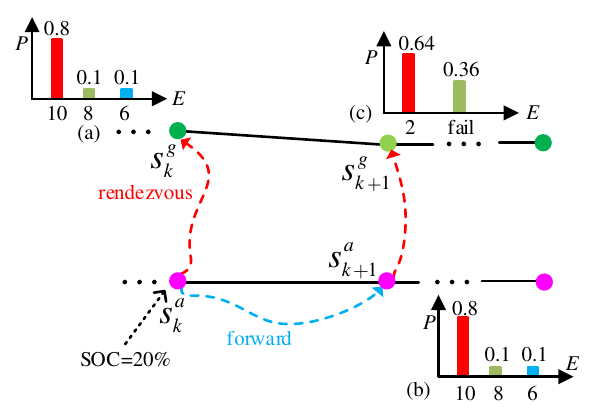}
    \caption{An illustrative example to explain the policy constructed with occupancy measures from LP\@. When the UAV is at state $s_{k}^a$ with State of Charge (SOC) equal 20\%, it has two actions: move forward (blue dashed line) to its next task node $s_{k+1}^a$ or rendezvous with the UGV (red dashed line) to recharge itself. The distribution of SOC after taking rendezvous is shown in (a) and that for forward action is shown in (b). (c) shows the distribution of SOC if the UAV first chooses forward action at $s_{k}^a$ and then chooses rendezvous $s_{k+1}^a$.  }
    \label{fig:different_risk_tolerance_policy}
\end{figure}

An illustrative example to explain the policy extracted from the solution of LP is given in Fig. \ref{fig:different_risk_tolerance_policy}. When the UAV reaches $s_{k}^a$, it has 20\% battery remaining. It can choose to rendezvous to replenish itself or move forward to its next task node $s_{k+1}^a$. 
If the UAV chooses to rendezvous, the distribution of the battery remaining at the rendezvous point is shown in    Fig. \ref{fig:different_risk_tolerance_policy}.(a). The distribution corresponding to forward action is given in Fig. \ref{fig:different_risk_tolerance_policy}.(b). If we set $\delta=0.01$, the policy constructed from  LP results in rendezvous with a probability 0.997 and forward with 0.003. 
Otherwise, if the UAV were to choose to move forward, it would have less than 10\% battery charge remaining with a probability of 0.2, which is not enough to ensure a rendezvous or forward action at $s_{k+1}^a$, suggesting  the failure probability is at least 0.2. By contrast, if we set the $\delta=0.5$, the policy becomes that rendezvous with a probability 0.504 and move forward with 0.496. One reason for such a policy is that with a probability of 0.8 the UAV will have 10\% battery remaining when it moves to $s_{k+1}^a$, where a rendezvous process with a shorter travel distance may be possible with a lower failure probability than the risk tolerance as shown in Fig.\ref{fig:different_risk_tolerance_policy}.(c).

\section{Experiments}\label{section:numerical simulation}

\begin{figure*}[t]
    \centering
    \subfloat[]{
    \includegraphics[width=0.25 \textwidth]{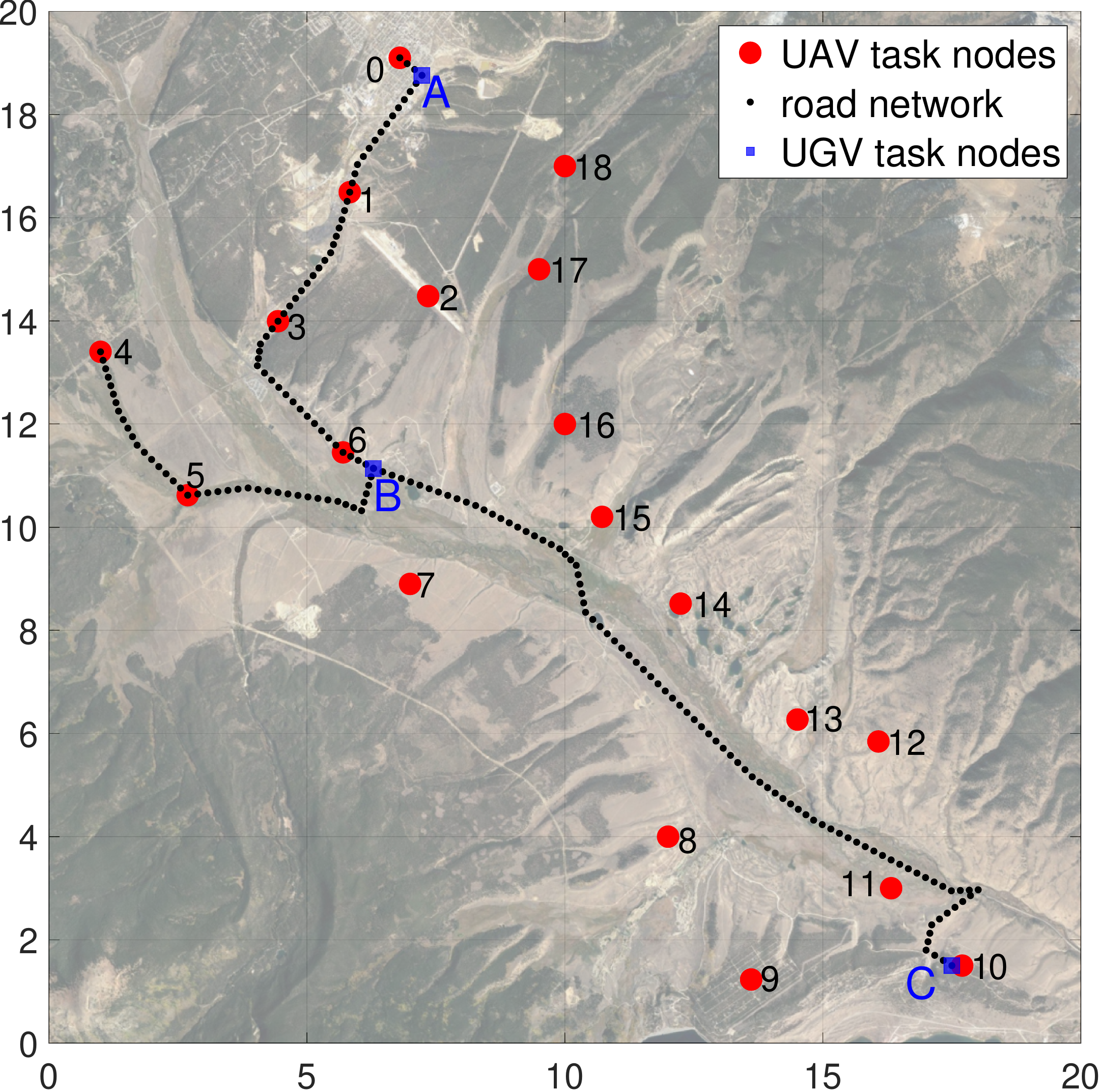}
    \label{fig:rendezvous_input}
    } 
    \subfloat[]{
    \includegraphics[width=0.28\textwidth]{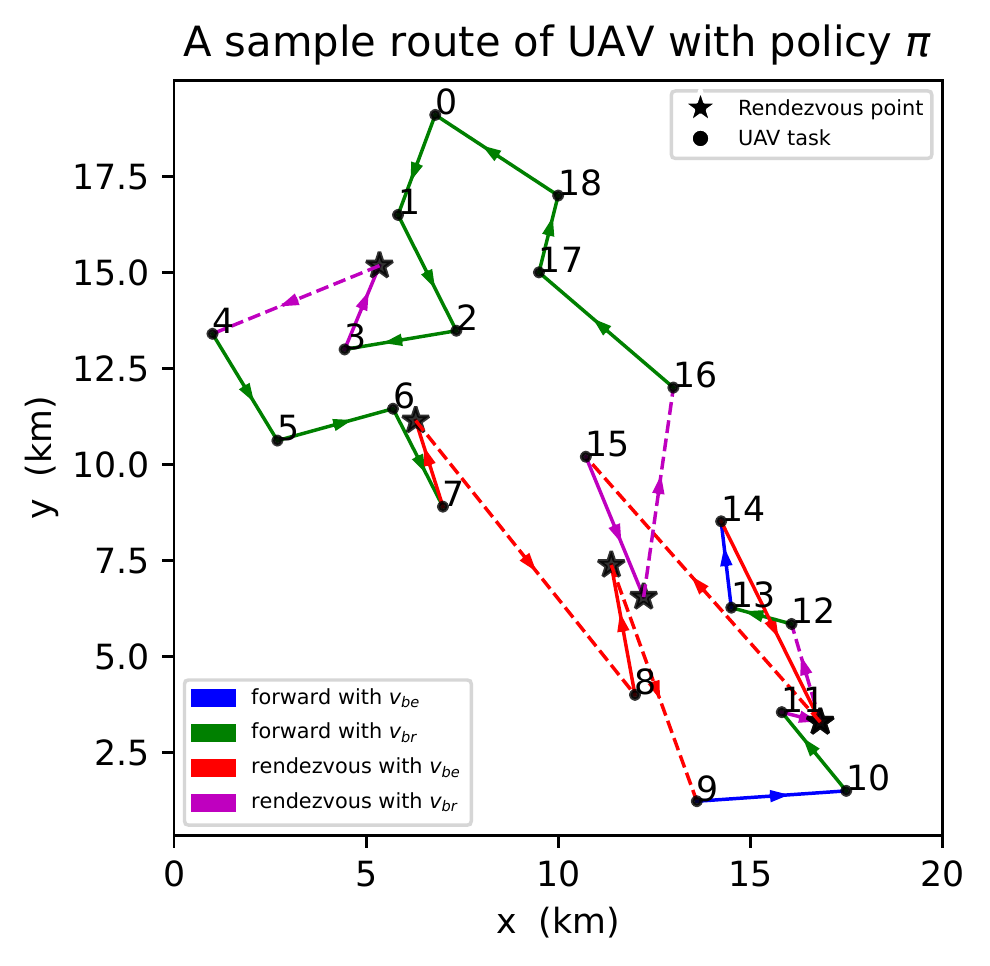}
    \label{fig:rendezvous_output_uav}
    }
    \subfloat[]{
    \includegraphics[width=0.28\textwidth]{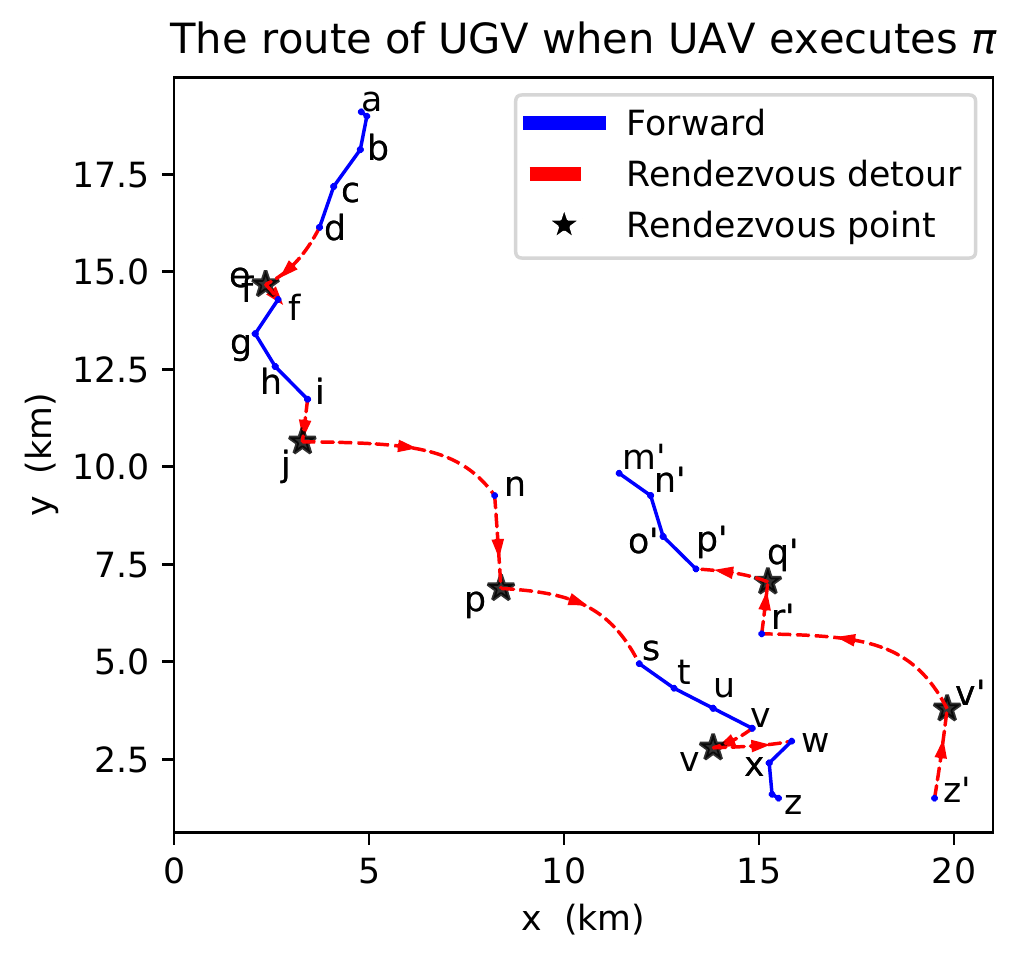}
    \label{fig:rendezvous_output_ugv}
    }
    \caption{
     A qualitative example to illustrate how UAV and UGV rendezvous with each other under the policy $\pi$ that is obtained by solving the CMDP\@. The risk tolerance is set to be $\delta=0.1$ in this case study. (a) The input of the risk-aware rendezvous problem. (b) One sample route of UAV when it executes the policy $\pi$. (c) The route of UGV corresponds to the route of UAV\@. Lower case letters with or without prime denote the same road node that is visited at different times}
    \label{fig:rendezvous_illustrative_example}
\end{figure*}
In this section, we first present a qualitative example to show what the input and output look like for our problem. Next, we study how system parameters (different risk tolerances) influence the rendezvous behaviors between the UAV and the UGV\@.  Then, we present quantitative results for the ISR application that motivates our research. Specifically, we will use Monte Carlo (MC) simulations to evaluate 
\begin{enumerate*}
    \item the satisfaction of the risk constraint for the policy constructed from LP;
    \item the effectiveness of the policy in minimizing the expected task duration;
    \item the risk tolerance-task duration Pareto curves.
\end{enumerate*}
Moreover, the running time of LP for CMDP is empirically evaluated. All experiments are conducted using Python 3.8 in a PC with i9-8950HK processor. LP is solved using Gurobi 9.5.0. 

\subsection{{Task route planner}} \label{subsection: UAV_task_planning}
{The task routes $\mathcal{T}_a$ and $\mathcal{T}_g$ used in Problem \ref{problem:risk-aware-rendezvous} can be either generated jointly by some existing task planners  \cite{manyam2019cooperative, yu2018algorithms} or can be generated by separately by different task planners. In our case study, the task for the UGV is to persistently monitor nodes A, B, and C (blue squares in Fig. \ref{fig:rendezvous_illustrative_example}). The task nodes for the UAV are red dots in Fig. \ref{fig:rendezvous_illustrative_example} and the task route (from node 0 to 18 and back to 0) is generated by a planner for Traveling Salesman Problem (TSP). }
\subsection{System Models}
\begin{table}[ht]
\centering
\caption{Coefficients for stochastic energy consumption model}
\begin{tabular}[t]{lcccccc}
\toprule
&$b_0$ &$b_1$ &$b_2$ &$b_3$ &$b_4$ &$b_5$\\
\midrule
{Value} &-88.77 &3.53 &-0.42 &0.043  &107.5  &-2.74\\
\bottomrule
\end{tabular}
\label{table:coefficient}
\end{table}
The UAV task and UGV tasks are from our ongoing project on intelligence, surveillance, and reconnaissance (ISR) as shown in Fig. \ref{fig:rendezvous_input}. In this project, we are interested in the case where $\delta=0.1$. UAV has about 240 KJ energy and its best range speed and best endurance speed are 14 m/s and 9.8 m/s respectively. UGV moves at 4.5 m/s. The rendezvous process will take 300 seconds. 

\begin{figure}
    \centering
    \includegraphics[width=0.28 \textwidth]{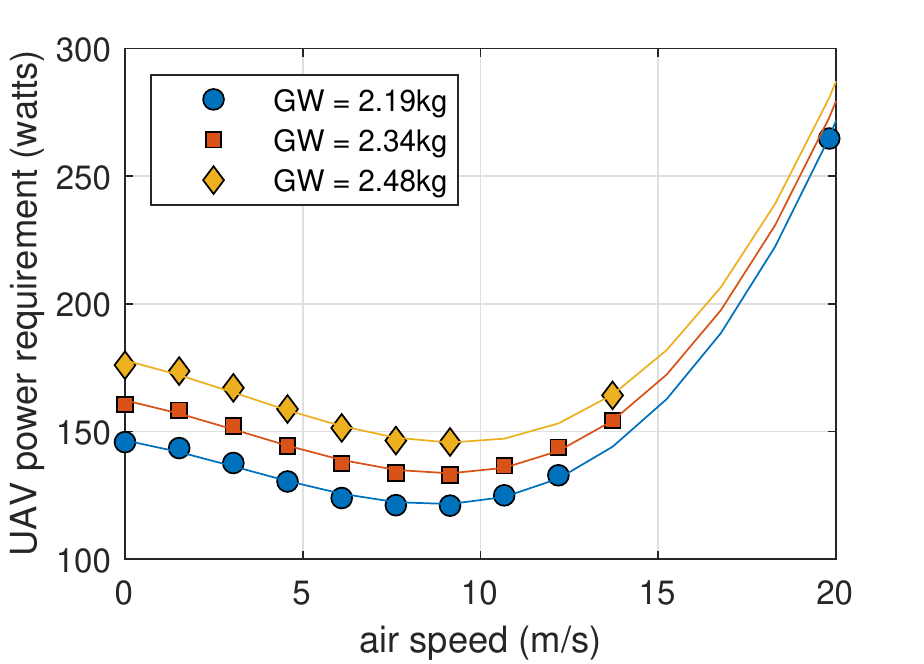}
    \label{fig:power_model_fit}
    \caption{
     Comparison of analytical data used to derive the polynomial regression fit model of UAV power requirement at three weights and across 11 airspeeds. }
    \label{fig:power_consumption}
\end{figure}

We consider two sources of stochasticity in the energy consumption model of UAVs: weight and wind velocity contribution to longitudinal steady airspeed.
The deterministic energy consumption model of the UAV is a polynomial fit constructed from analytical aircraft modeling data, given as
\begin{equation}
    P(\bm{v_\infty}) = b_0 + b_1 \bm{v_\infty} + b_2 \bm{v_\infty}^2 + b_3 \bm{v_\infty}^3 + b_4 \bm{w} + b_5\bm{v_\infty} \bm{w}, 
\end{equation}
where $b_0$ to $b_5$ are coefficients, and their experimental values are listed in Table \ref{table:coefficient}.
Figure~\ref{fig:power_consumption} shows the agreement between the polynomial regression fit model and the analytical data that it was derived from.

Weight is randomly selected following a normal distribution with a mean of 2.3 kg and a standard deviation of 0.05 kg, $\bm{w} \sim \mathcal{N}(\mu_{\bm{w}}, \sigma^2_{\bm{w}})$.
Vehicle airspeed, $v_\infty$, is the sum of the vehicle ground speed, $v$, and the component of the wind velocity that is parallel to the vehicle ground speed, ignoring sideslip angle and lateral wind components.
\begin{equation}
    v_\infty = \lvert \Bar{v_g} + \rm{cos}(-\psi)\bm{\xi}_{a, b} \rvert 
\end{equation}
The longitudinal wind speed contribution is derived from two random parameters; wind speed, and wind direction.
Wind speed is modeled using the Weibull probability distribution model of wind speed distribution, $\bm{\xi}_{a, b}$, with a characteristic velocity $a=1.5$ m/s and a shape parameter $b=3$. This is representative of a fairly mild steady wind near ground level.
Wind direction $\psi$ is the heading direction of the wind, and is uniformly randomly selected on a range of $[0,360)$ degrees. 
\begin{figure*}
    \centering
    \subfloat[Low risk tolerance ($\delta=0.01$)]{
    \includegraphics[width=0.26\textwidth]{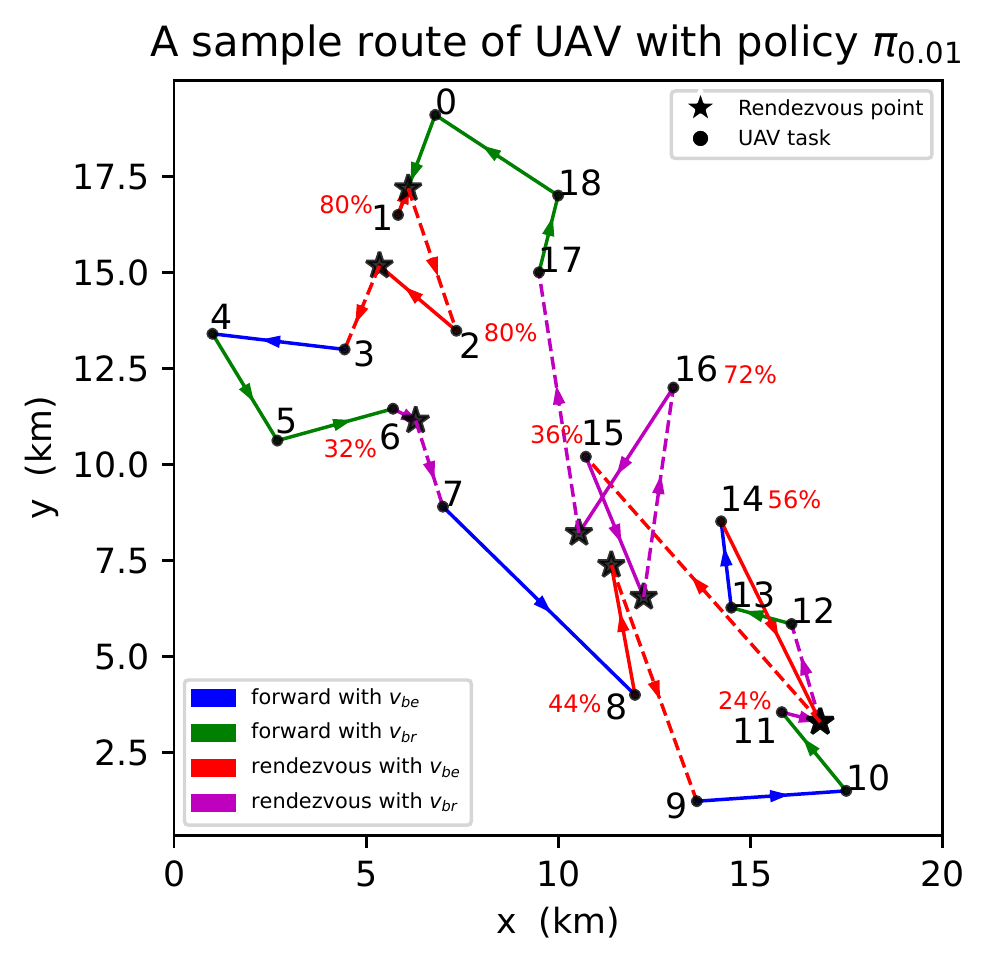}
    \label{fig:risk_example0.01_UAV}
    }
    \subfloat[Medium risk tolerance ($\delta=0.2$)]{
    \includegraphics[width=0.26\textwidth]{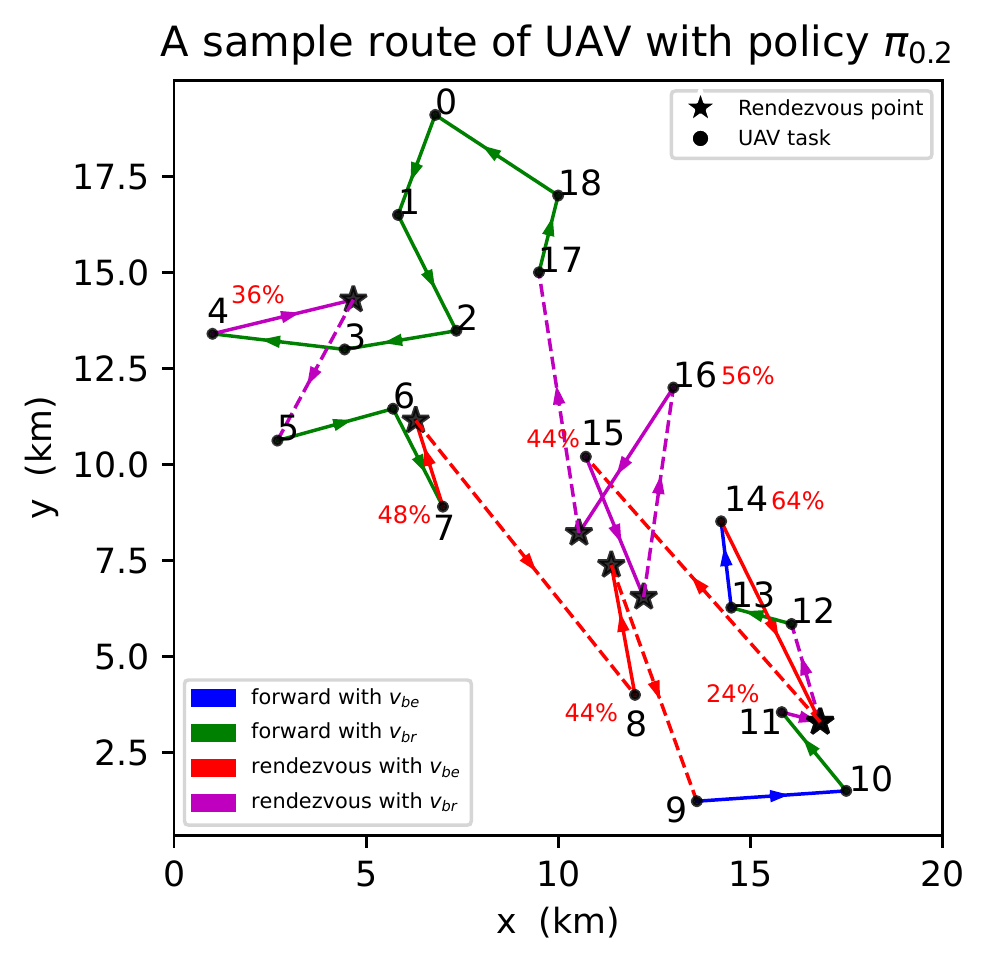}
    \label{fig:risk_example0.2_UAV}
    }
    \subfloat[High risk tolerance ($\delta=0.5$)]{
    \includegraphics[width=0.26\textwidth]{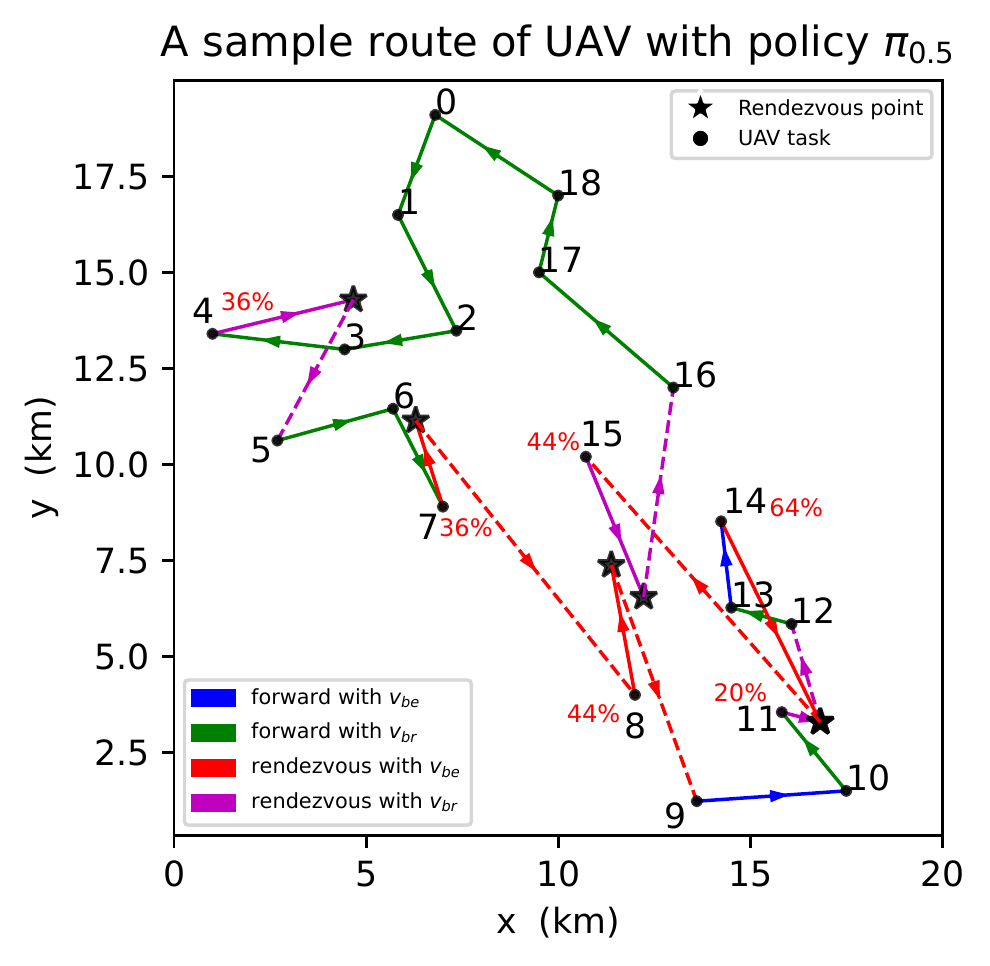}
    \label{fig:risk_example0.5_UAV}
    } \\
    \subfloat[UGV route corresponding to Fig. \ref{fig:risk_example0.01_UAV}]{
    \includegraphics[width=0.26\textwidth]{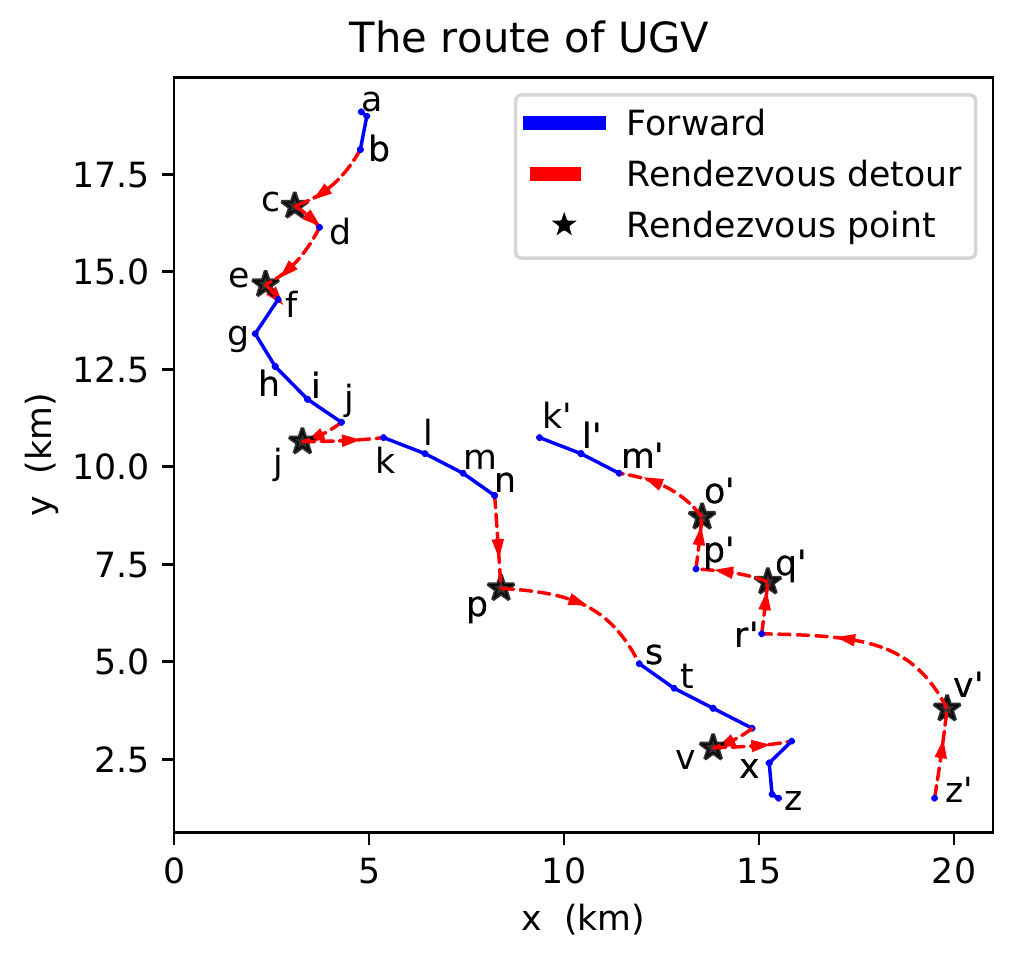}
    \label{fig:risk_example0.01_UGV}
    }
    \subfloat[UGV route corresponding to Fig. \ref{fig:risk_example0.2_UAV}]{
    \includegraphics[width=0.26\textwidth]{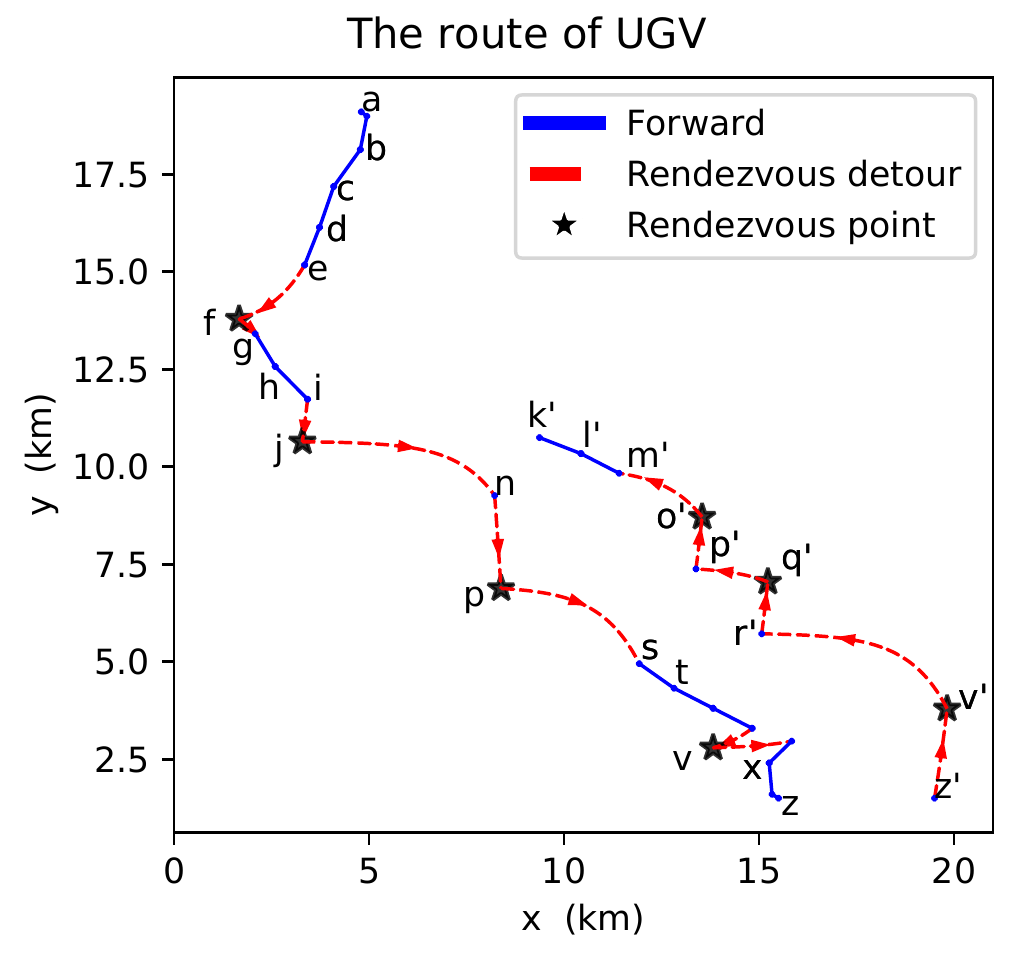}
    \label{fig:risk_example0.2_UGV}
    }
    \subfloat[UGV route corresponding to Fig. \ref{fig:risk_example0.5_UAV}]{
    \includegraphics[width=0.26\textwidth]{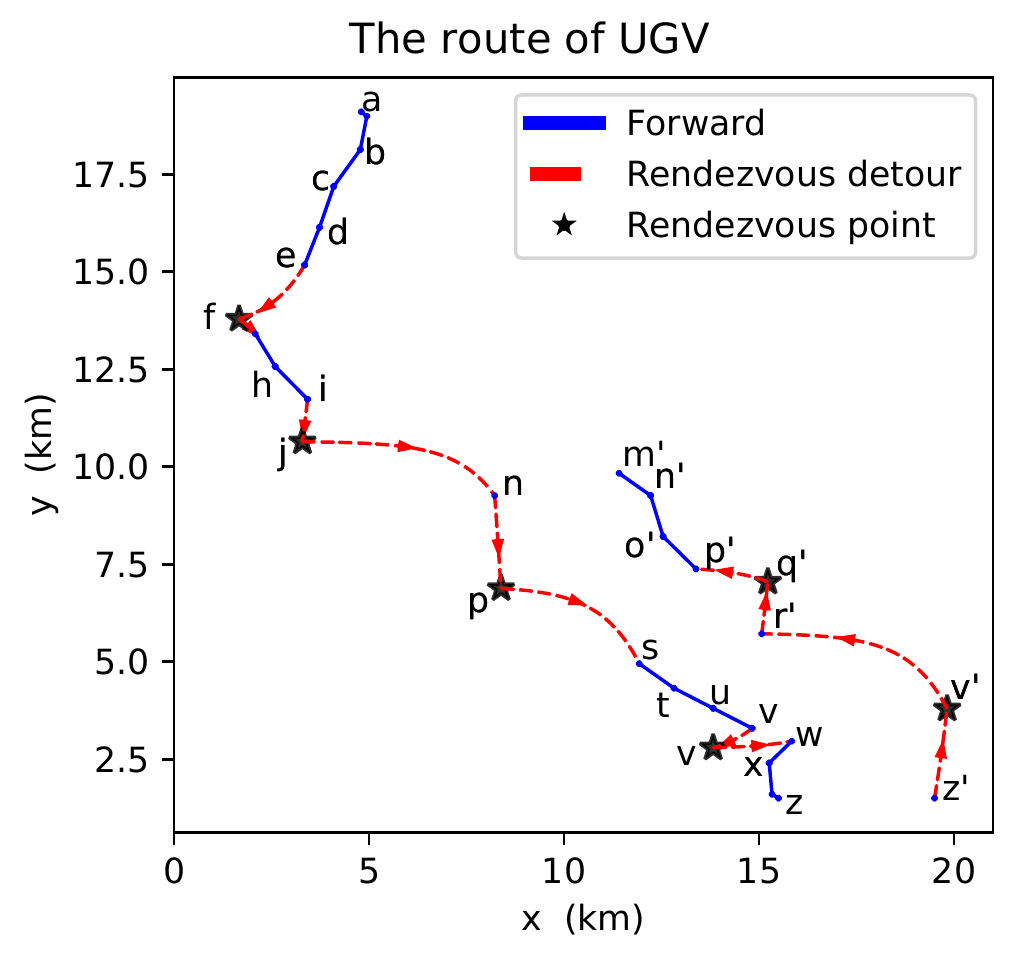}
    \label{fig:risk_example0.5_UGV}
    }\\
     \subfloat[Quantitative results.]{
        %\caption{Quantitative results for results }
        \begin{tabular}[t]{lccc}
            \toprule
            UAV data &$\delta=0.01$ &$\delta=0.2$ &$\delta=0.5$ \\
            \midrule
            {Empirical failure rate} &0.00975 &0.201 &0.497 \\
            Average route travel time  &11330 s &{10418} s &10404 s \\
            route travel time overhead &160.2\% &139.2 \% &138.9\% \\
            Average route travel distance &96.8 km &97.8 km &97.6 km \\
            Travel distance overhead &58.2\% &60.5\% &60.3\% \\
            Average \# of rendezvous &8.5 &6.4 &6.3 \\
            \bottomrule
        \end{tabular}
    \label{table:risk_example_table}    
    } 
    
    \caption{ How different risk thresholds influence the rendezvous behaviors. UAV route time with the best range speed is 4354 s and the route distance is 61.0 km. (a) UAV is very risk-averse to failures with a risk threshold equal to 0.01. (b) UAV is less risk-averse to failures with a risk threshold equal to 0.2. (c) UAV is neutral to the failures with a risk threshold equal to 0.5.}
    \label{fig:risk_averse_example}
\end{figure*}

\subsection{Simulation Results}
An illustrative example of the input and the output of the problem considered is shown in Fig. \ref{fig:rendezvous_illustrative_example}. The input of the problem is shown in Fig. \ref{fig:rendezvous_input}, which consists of UAV task nodes (red dots), UGV task nodes (blue square), and road network (black nodes). Fig. \ref{fig:rendezvous_output_uav} and \ref{fig:rendezvous_output_ugv} shows one sample route of UAV and UGV respectively when the system executes the policy computed by LP. UAV's route starts from node 0. When the UAV reaches node 4, it will choose to rendezvous with UGV using the best range speed in a rendezvous point, which is denoted as a star, and then go to its next task node 5. Similarly, the UAV will rendezvous with the UGV when it reaches nodes 7, 8, 11, 14, and 15. The corresponding route of UGV is presented in Fig.  \ref{fig:rendezvous_output_ugv}. 
 
Next, we show how different risk tolerances influence the rendezvous behaviors under our CMDP formulation. In these experiments, we set the risk tolerance $\delta$ to be 0.01, 0.2, and 0.5. {Results shown in Fig. \ref{fig:risk_averse_example}} includes sample routes for the UAV and the UGV and statistical data of the policies. Fig. \ref{fig:risk_example0.01_UAV}, \ref{fig:risk_example0.2_UAV}, and \ref{fig:risk_example0.5_UAV} are sample routes for the UAV when it executes the policy. Fig. \ref{fig:risk_example0.01_UGV}, \ref{fig:risk_example0.2_UGV}, and \ref{fig:risk_example0.5_UGV} are corresponding routes of the UGV\@. The rendezvous point is denoted as a star. The SOC is annotated in red text close to the task node at which the UAV decides to rendezvous. Some statistical data are summarized in table \ref{table:risk_example_table}. In general, we observe that when the risk tolerance is set to be small, the UAV tends to rendezvous more often, and the average route travel time is higher. Here the average route travel time is computed by considering only trials in which the UAV finishes its task route. By contrast, as the risk tolerance is relaxed to a larger value, the average route travel time will decrease, which comes at the cost of a high failure probability. 

We also conducted several quantitative experiments to validate our formulation. 
  The first experiment is to use MC simulation to check whether the failure probability is upper bounded by the set risk tolerance of 0.1. 
We use FR to denote the empirical failure rate.  As can be seen in Table \ref{table:MC_convergence}, as MC increases, the empirical failure rate is 
close to and below the theoretical PF 0.1. In the following experiments, we will use $N_{\scaleto{MC}{3pt}}=2000$ for simulation.

\begin{figure}
    \centering
    \includegraphics[width=0.35\textwidth]{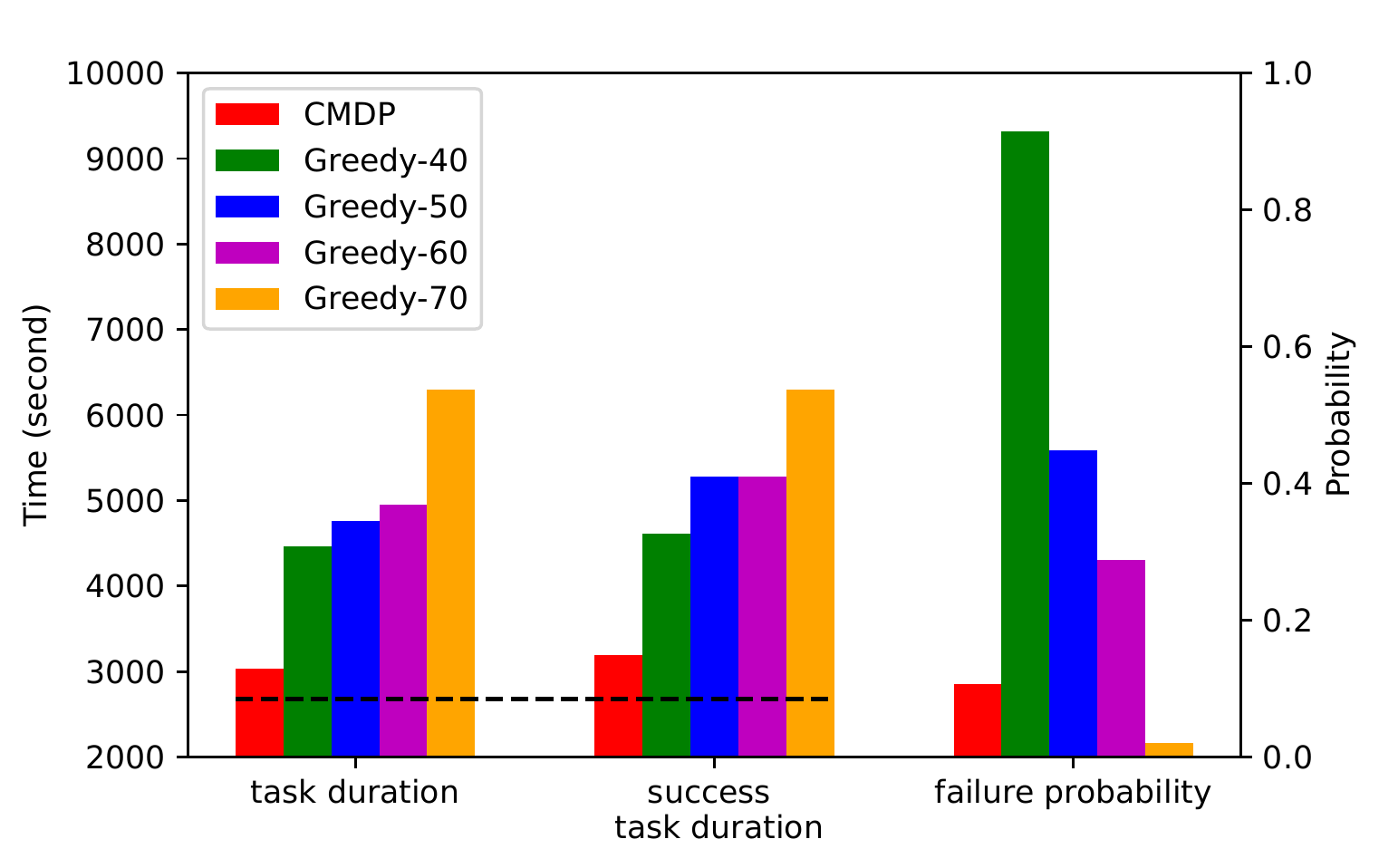}
    \caption{Results comparisons for the CMDP policy and the greedy policies with $\delta=0.1$. The dashed black line represents the task duration if the UAV moves with the best range speed without considering battery limitation. }
    \label{fig:comparison}
\end{figure}
\begin{table}[t]
\centering
\caption{Empirical evaluation of failure probability. $\delta =0.1$. }
\begin{tabular}[t]{lcccc}
\toprule
$N_{\scaleto{MC}{3pt}}$ (\# of MC trials)&$500$ &$1000$ &$3000$ &$5000$\\
\midrule
{Failure rate} &0.108 &0.105 &0.099 &0.097\\
\bottomrule
\end{tabular}
\label{table:MC_convergence}
\end{table}
\begin{figure}
    \centering
    \includegraphics[scale=0.40]{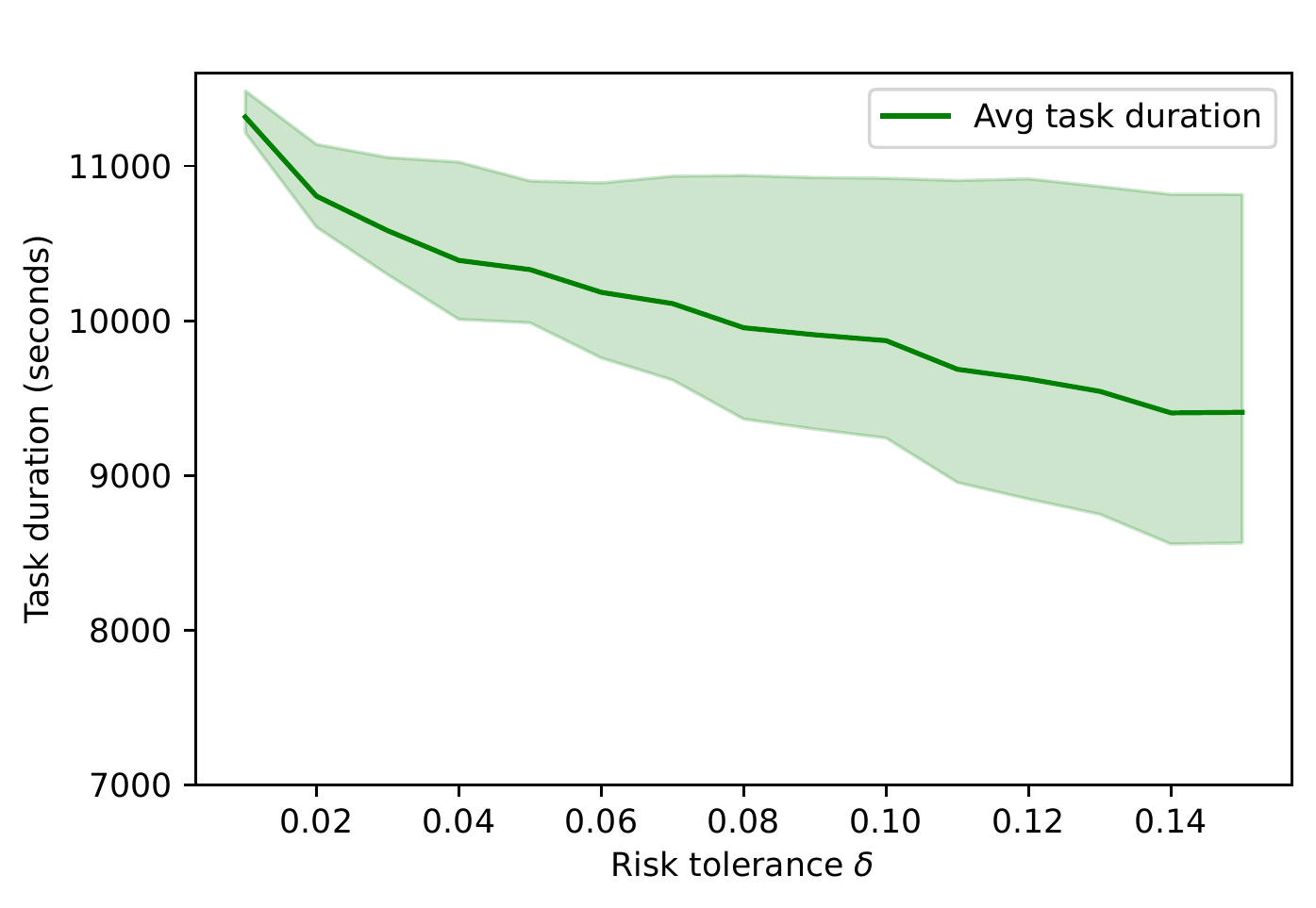}
    \caption{Risk tolerance vs task duration Pareto Curve.  }
    \label{fig:pareto_curve}
\end{figure}
To validate that the  policy  constructed  from  LP can minimize the expected travel distance. We compare our policy with a greedy baseline. The greedy policy is set as: always flies with the best range speed and chooses to rendezvous when state-of-charge drops below a set value. {What we observe in experiments is that when the route of a UAV is long, for example, there are more than 15 nodes, the probability of finishing the route is close to zero for the greedy baseline no matter what threshold we set. To have a more informative comparison, we use only nodes 0 to 11 for the task route in the following experiments.} We consider four-set values $40\%$,  $50\%$, $60\%$, and $70\%$, and the corresponding policies are denoted as \textit{Greedy-40}, \textit{Greedy-50}, \textit{Greedy-60}, \textit{Greedy-70}. As shown in Fig.\ref{fig:comparison}, our policy can guarantee success probability above the set value of 0.9 and the expected travel time of UAV is shorter compared to the baseline. Though the baseline can achieve a higher success probability in some cases as shown in Fig.\ref{fig:comparison}, its expected task duration is still longer than our policy.  

The empirical Pareto curve for risk tolerance and the task duration is shown in Fig. \ref{fig:pareto_curve}. The green curve is the mean value and the shaded area is formed using one standard deviation from the mean. When the risk level is set to be a higher value, the UAV will tend to make more risky decisions, leading to a lower travel time at the cost of a higher failure probability. 

\iffalse
\begin{table}[ht]
\centering
\caption{The gap between task duration and conditional task duration.}
\begin{tabular}[t]{lccccc}
\toprule
&$\delta = 0.01$ &$\delta = 0.03$ &$\delta = 0.05$ &$\delta = 0.08$ &$\delta = 0.1$\\
\midrule
Error &0.18\% &1.95\% &3.01\% &5.6\% &6.18\%\\
\bottomrule
\end{tabular}
\label{table:objective_gap}
\end{table}

\subsubsection*{Discussion} 
In this paper, the objective for CMDP is to minimize the task duration the same as those in the related work \cite{chow2015trading}. One alternative objective is to minimize the task duration that only considering the route that the UAV successfully finishes its task route. However, as suggested \cite{chow2015trading}, when the risk level is small, for example, $\delta \leq 0.1$, the solution gap between these two formulations is small. We empirically compare these two formulations. As shown in Table \ref{table:objective_gap}, the gap is only about $6\%$ when $\delta \leq 0.1$, which is the case we are interested in for applications. {One way to remove this gap for our application is to modify the CMDP formulation. Instead of letting the state transit to the terminal state when the UAV runs out of battery, we allow the UAV and UGV to transit back to their tasks with an additional cost added to the main objective, which in practice represents the time consumed to do failure recovery for the UAV\@. Another direction to remove the gap would be to find new algorithms that can minimize the expected length of only those successful routes. }
\fi

The running time for the proposed routing problem consists of three parts. The first part is devoted to extracting transition information for LP\@. The second part is about constructing an LP model with Gurobi and the last part is for solving the LP\@. In our case study, there are about 54000 states and it takes about 6 min to extract transition information, 9 minutes to create an LP model, and about 1 second to solve the LP\@.

\iffalse        
$$
\begin{aligned}
\mathbb{E} \left [ \rm{task~duration} \right ] = \mathbb{E} \left [ \rm{task~duration}  \mid \rm{success} \right]P(\rm{success}) \\
+ \mathbb{E} \left [ \rm{task~duration}  \mid \rm{failure} \right]P(\rm{failure}) 
\end{aligned}
$$

$$
\begin{aligned}
 \frac{\mathbb{E} \left [ \rm{task~duration} \right ]}{P(\rm{success})}= \mathbb{E} \left [ \rm{task~duration}  \mid \rm{success} \right] \\
+ \mathbb{E} \left [ \rm{task~duration}  \mid \rm{failure} \right]\frac{P(\rm{failure})}{P(\rm{success})} \\
\leq \mathbb{E} \left [ \rm{task~duration}  \mid \rm{success} \right]
+ \\ 
\mathbb{E} \left [ \rm{task~duration}  \mid \rm{success} \right]\frac{P(\rm{failure})}{P(\rm{success})}
\end{aligned}
$$

$$
\begin{aligned}
 \frac{\frac{\mathbb{E} \left [ \rm{task~duration} \right ]}{P(\rm{success})} - \mathbb{E} \left [ \rm{task~duration}  \mid \rm{success} \right]}{\mathbb{E} \left [ \rm{task~duration}  \mid \rm{success} \right]} \\
\leq 
\frac{P(\rm{failure})}{P(\rm{success})}
\end{aligned}
$$
\fi

\section{Conclusion}
In this paper, we study a variant of the cooperative aerial-ground routing problem with energy chance constraint on the UAV\@. We formulate the problem as a CMDP and use LP to find the optimal policy for the CMDP\@. We validate our formulation and the solution in one ISR application. In future work, one direction we will explore is to extend one UAV one UGV routing problems to multiple UAVs and UGVs and consider the distributed solution for such types of risk-aware routing problems. 
\label{section:conclusion}

\bibliographystyle{IEEEtran}
\bibliography{IEEEabrv,CDC2022}

\end{document}